\documentclass{article}

\usepackage[nonatbib,preprint]{neurips_2023}

\usepackage[utf8]{inputenc} 
\usepackage[T1]{fontenc}    
\usepackage{hyperref}       
\usepackage{url}            
\usepackage{booktabs}       
\usepackage{amsfonts}       
\usepackage{nicefrac}       
\usepackage{microtype}      
\usepackage{xcolor}         
\usepackage{cite}
\usepackage{graphicx}
\usepackage{subfigure}
\usepackage{multirow}
\usepackage{amsmath, amsthm, amssymb}
\usepackage{makecell}
\usepackage{tablefootnote}

\title{TigerBot: An Open Multilingual Multitask LLM}

\author{
  Ye~Chen\\
  Tiger Research\\
  Shanghai, China \\
  \texttt{yechen@tigerbot.com} \\
   \And
  Wei~Cai \\
  Tiger Research \\
  Shanghai, China \\
  \texttt{wei.cai@tigerbot.com} \\
  \And
  Liangmin~Wu \\
  Tiger Research\\
  Shanghai, China \\
  \texttt{liangmin.wu@tigerbot.com} \\
  \And
  Xiaowei~Li \\
  Tiger Research \\
  Shanghai, China \\
  \texttt{xiaowei.li@tigerbot.com} \\
  \And
  Zhanxuan~Xin \\
  Tiger Research \\
  Shanghai, China \\
  \texttt{zhanxuan.xin@tigerbot.com} \\
  \And
  Cong~Fu \\
  Tiger Research \\
  Shanghai, China \\
  \texttt{cong.fu@tigerbot.com} \\
}

\begin{document}

\maketitle

\begin{abstract}
We release and introduce the TigerBot family of large language models (LLMs)~\footnote{web: \url{https://www.tigerbot.com/chat}; github: \url{https://github.com/TigerResearch/TigerBot}}, consisting of base and chat models, sized from 7, 13, 70 and 180 billion parameters. We develop our models embarking from Llama-2 and BLOOM, and push the boundary further in data, training algorithm, infrastructure, and application tools. Our models yield meaningful performance gain over SOTA open-source models, e.g., Llama-2, specifically 6\% gain in English and 20\% gain in Chinese. TigerBot model family also achieves leading performance in major academic and industrial benchmarks and leaderboards~\footnote{As of this writing, TigerBot ranked top-tier open-source models in~\href{https://opencompass.org.cn/leaderboard-llm}{OpenCompass LLM Leaderboard }, and \href{https://github.com/jeinlee1991/chinese-llm-benchmark}{CLiB Chinese LLM benchmark leaderboard}.}. We believe that TigerBot represents just a snapshot of lightning-fast progression in LLM open-source community. Therefore, we are thrilled to give back by publicly releasing our models and reporting our approach behind, with additional emphases on building SOTA LLMs in a democratized way and making LLMs of use in real-world applications.
\end{abstract}

\section{Introduction}
Large language models (LLMs) has shown unprecedented promise in a wide range of tasks. Since the phenomenal launch of ChatGPT, there have been breathtaking development in the community, mainly following three themes:
\begin{enumerate}
	\item Fundamental capabilities, pushed forward by both proprietary models (GPT~\cite{Ouyang:2022aa,Brown:2020aa}, BARD~\cite{Pichai:2023}, Claude~\cite{Claude:2023}) and  open-source models (BLOOM~\cite{LeScao:2022aa}, Llama~\cite{Touvron:2023ab,Touvron:2023aa}).
	\item Computational economics, from data collection (e.g. Alpaca~\cite{Taori:2023}), training (e.g. LoRA~\cite{Hu:2021aa}), quantization (ExLlama~\cite{Turboderp:2023}), and inference (e.g. TGI~\cite{Huggingface:2023ab} and TensorRT~\cite{Nvidia:2023}).
	\item Application readiness, from APIs, plug-ins, function calling and agents, retrieval-augmented generation (RAG), long context window, to recently multimodality and role-playing.	
\end{enumerate}

The mainstream approach to building LLMs has been pretraining decoder-only transformers~\cite{Vaswani:2017aa} on an extensive corpus of unsupervised textual data, followed by alignment with human preferences with labelled demonstration or comparison data, using supervised fine-tuning (SFT) or reinforcement learning with human feedback (RLHF). We have followed the same methodology, albeit made the following contributions:
\begin{enumerate}
	\item A new training data mix with thorough experimental assessment and cleaning.
	\item A stack of novel algorithmic and infrastructural implementations to make our models state-of-the-art (SOTA) in both performance and computational efficiency.
	\item A thorough description of our implementations and observations from the field, in deploying our models to real-world applications, which help us prioritize research endeavors.
\end{enumerate}
Besides achieving superior fundamental capabilities, we are dedicated to democratizing LLM development. To the best of our knowledge, TigerBot only incurs the least amount of computational costs (less than two million dollars over the time period April--December, 2023) and carbon footprint to produce probably one of the most comprehensive model families (from 7B to 180B, base and chat, with full stack of tools). This can only happen with an open-source spirit, hence we contribute a detailed elaboration on our methodology and experiences by the same token. Furthermore, we have taken measures to ensure the safety of our models.

\section{TigerBot models}
\begin{figure}[h]
\centerline{
\subfigure[Training loss for Tigerbot-70b etc.]{\includegraphics[width=0.4\linewidth]{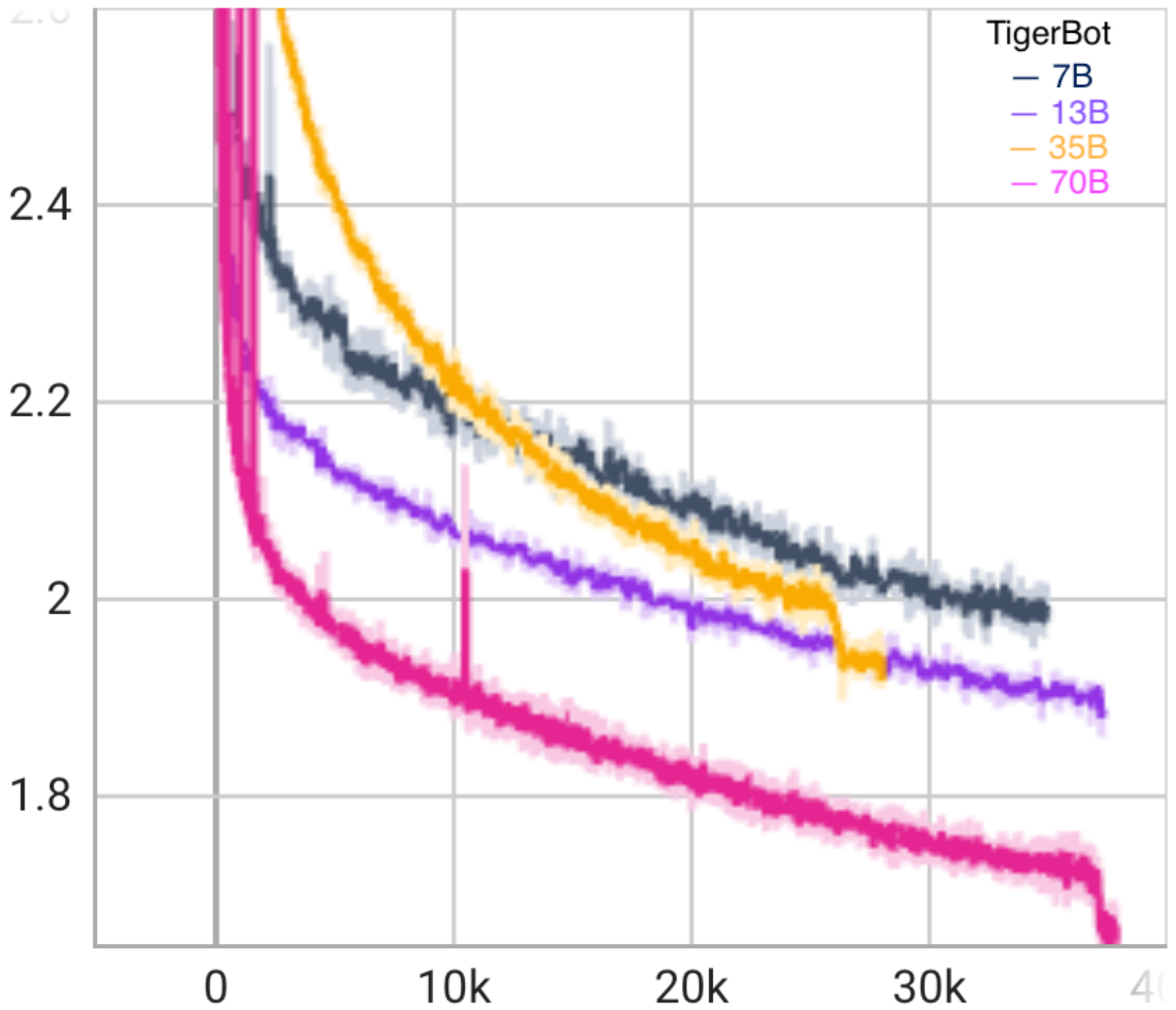}
\label{fig-train-loss-70b-etc}}
\subfigure[Training loss for Tigerbot-180b]{\includegraphics[width=0.41\linewidth]{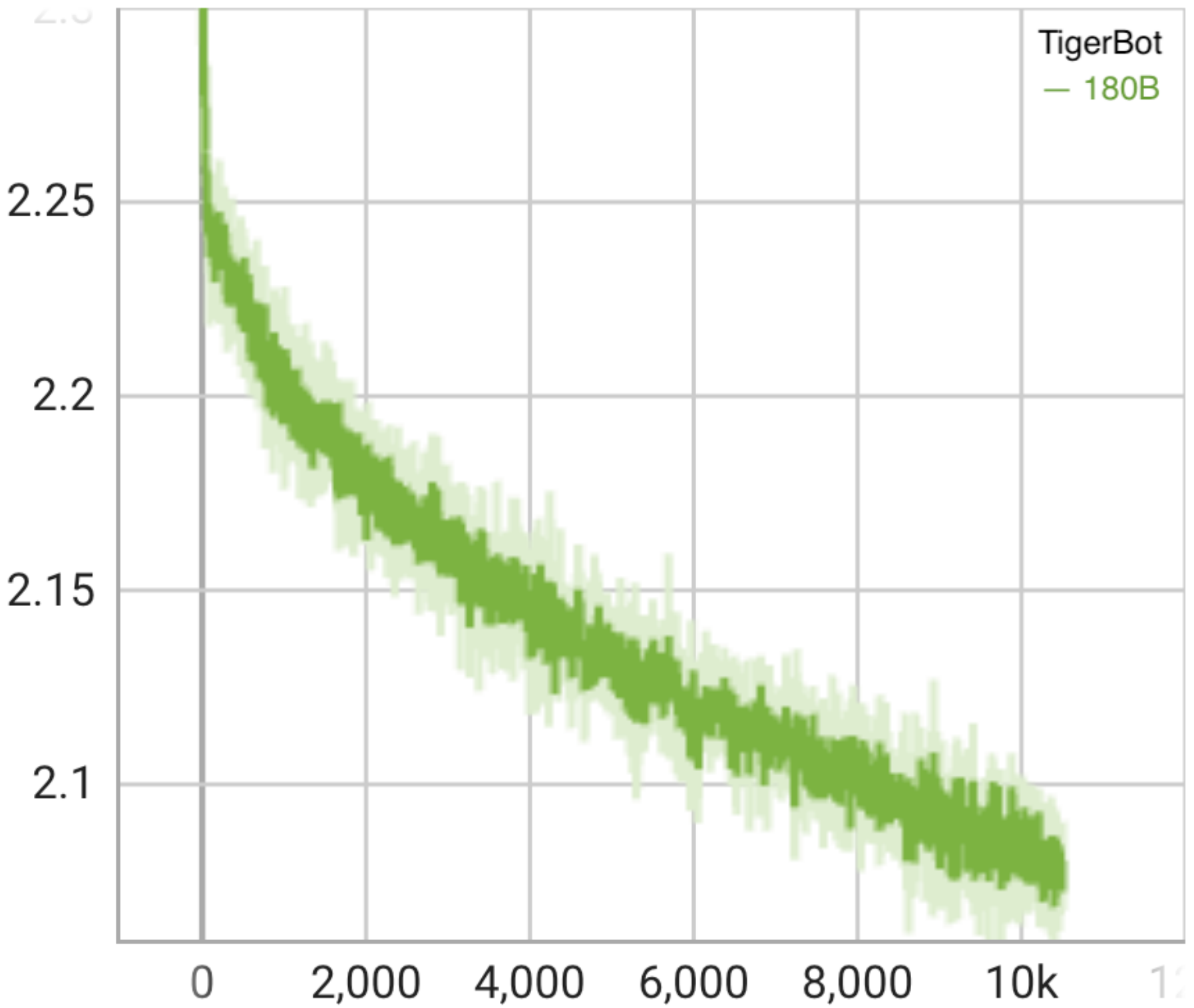}
\label{fig-train-loss-180b}}
}
\caption{Training loss for Tigerbot models}
\label{fig-train-loss}
\end{figure}

We are open-source releasing our models for free research and commercial use~\footnote{TigerBot is released under Apache-2.0 license. However, since we continual pretrained from Llama-2 (7, 13, 70B) and BLOOM (180B), one shall consult with their open-source licenses respectively.}, as summarized in Table~\ref{tab-models}, along with a suite of developer tools. Figure~\ref{fig-train-loss} shows the training loss for pretraining.

\begin{table}[h]
  \caption{Tigerbot model family}
  \label{tab-models}
  \centering
  \begin{tabular}{clclclclclc|c}
    \toprule
    Model & Base & Chat & API & Plug-in & Multi-modal & Context-length\\
    \midrule
    7B & \checkmark & \checkmark & chat, fine-tune & search & image out & 2k\\
    13B & \checkmark & \checkmark & chat, fine-tune & search, doc & image in/out & 32k\\
    70B & \checkmark & \checkmark & chat, fine-tune & search, doc & image in/out & 32k\\
    180B & \checkmark & \checkmark & chat & search, doc & image in/out & 2k\\
    \bottomrule
  \end{tabular}
\end{table}

Base model is a manifestation of world knowledge, serving as a foundation for downstream fine-tuning and applications. Chat model is fine-tuned to perform general-purpose tasks such as chat, question-answering (QA), generation, and so forth. API is a quick way to tap into TigerBot SOTA model capabilities from cloud with few lines of codes. Moreover, plug-in's allow developers and users to leverage the entire internet through modern search engines (search), or their own proprietary knowledge base (document).

\subsection{Training data}\label{sec-data}
\begin{figure}[h]
\centering
\includegraphics[width=1\linewidth]{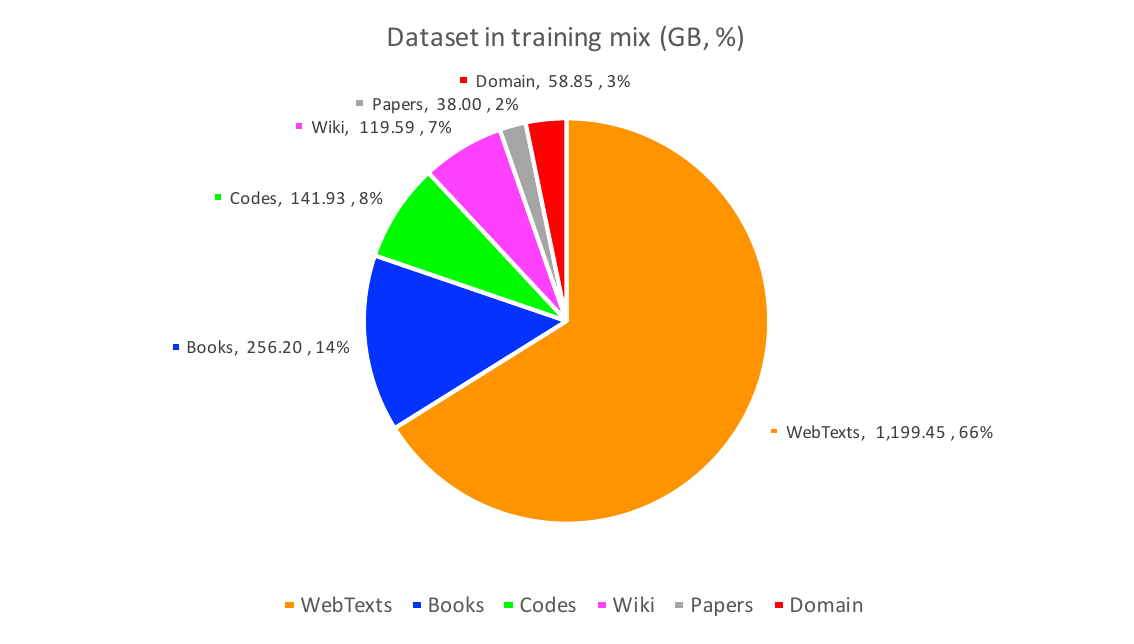}
\caption{Training data proportions}
\label{fig-train-mix}
\end{figure}

In the pretraining stage, our training data consists of about 500 billion tokens, or 1.8TB plaintext data, which in turn was cleaned, deduped and down-sampled from 5.6TB data. Our data are chosen from 25 public and proprietary datasets, based on the following design considerations: (1) good quality, in terms of factuality, diversity, and format, based on our months' iterations and user feedback; (2) multilingual coverage, especially Chinese (e.g., WuDao and WanJuan corpus) and major eastern asian languages besides English (with zh:en roughly 5:5); and (3) multitask coverage, such as web (C4 and RefinedWeb), books (BookCorpus and Clibrary), wikipedia, codes (GitHub and Stack Overflow), academics (arXiv), and domain data (e.g., legal and patent). Table~\ref{tab-train-data} shows our training data mixture and their sources, and Figure~\ref{fig-train-mix} illustrates proportions of datasets.

\begin{table}[h]
  \caption{Tigerbot training data}
  \label{tab-train-data}
  \centering
  \begin{tabular}{clclclc|c}
    \toprule
    \multicolumn{2}{c}{Dataset} & Size (GB) & Tokens (B) & Source \\
    \midrule
     \multirow{3}{*}{Books} & en-books & 100.00 & 25.06 & \href{https://github.com/soskek/bookcorpus}{BookCorpus} \\
     & zh-books & 154.00 & 39.23 & \href{https://clibrary.cn/}{Clibrary} \\
     & zh-textbook & 2.20 & 0.74 & \href{https://github.com/opendatalab/WanJuan1.0}{WanJuan} \\
     \midrule
     \multirow{5}{*}{WebTexts} & en-c4 & 80.30 & 19.92 & \href{https://huggingface.co/datasets/c4}{C4} \\
     & en-refinedweb & 345.15 & 86.80 & \href{https://huggingface.co/datasets/tiiuae/falcon-refinedweb}{RefinedWeb} \\
     & en-webtext & 39.00 & 10.14 & \href{https://huggingface.co/datasets/openwebtext}{OpenWebText} \\
     & zh-news & 121.00 & 27.38  & Tigerbot and \href{https://github.com/opendatalab/WanJuan1.0}{WanJuan} \\
     & zh-webtext & 614.00 & 147.59 & \href{https://github.com/BAAI-WuDao/Data}{WuDao} and \href{https://github.com/opendatalab/WanJuan1.0}{WanJuan} \\
    \midrule
    Papers	& en-arxiv	 & 38.00 & 12.52 & \href{https://arxiv.org/}{arXiv} \\
    \midrule
    \multirow{2}{*}{Codes} & en-github	 & 117.13 & 42.84 & \href{https://huggingface.co/datasets/codeparrot/github-code}{Github} \\
    & en-stackoverflow & 24.80 & 7.97 & \href{https://archive.org/download/stackexchange/}{Stack Overflow} \\ 
    \midrule
    \multirow{5}{*}{Wiki} & en-wiki & 21.00 & 6.68 & \href{https://dumps.wikimedia.org/enwiki/}{English wikipedia} \\
    & zh-wiki & 2.79 & 1.72 & \href{https://dumps.wikimedia.org/zhwiki/}{Chinese wikipedia} \\
    & zh-baike & 87.50 & 23.00 & Tigerbot and \href{https://github.com/BAAI-WuDao/Data}{WuDao} \\
    & ja-wiki & 6.80 & 2.00 & \href{https://dumps.wikimedia.org/jawiki/}{Japanese wikipedia} \\
    & ko-wiki & 1.50 & 0.53 & \href{https://dumps.wikimedia.org/kowiki/}{Korean wikipedia} \\
    \midrule
     \multirow{5}{*}{Domain} & en-stackexchange & 6.80 & 1.91 & \href{https://archive.org/download/stackexchange/}{Stack Exchange} \\
     & zh-law & 35.03 & 9.42 & Tigerbot and \href{https://github.com/opendatalab/WanJuan1.0}{WanJuan} \\
     & zh-patent & 17.00& 4.66 & \href{https://github.com/opendatalab/WanJuan1.0}{WanJuan} \\
     & zh-sentiment & 0.02 & 0.01 & \href{https://huggingface.co/datasets/sepidmnorozy/Cantonese_sentiment}{Cantonese sentiment} \\
     \midrule
    \multicolumn{2}{c}{\textbf{Total}} & \textbf{1,814.02} & \textbf{470.12}~\tablefootnote{later adding 5\% holistic training data to make $\sim$500B tokens.} && \\
    \bottomrule
  \end{tabular}
\end{table}
     
In the alignment learning stage, we collected 5 million instruction completion data for SFT, and 15k human annotated comparison data for RLHF, where 10k examples for rejection sampling and 5k for direct preference optimization (DPO). Our fine-tuning training data was chosen from about 30 open and proprietary datasets, and months of human annotation from real user prompts. The data mix is designed to represent a broad range of general-purpose tasks (e.g., generation, QA, and brainstorming), following a similar category in~\cite{Ouyang:2022aa} and we further expand to about 100 subcategories (e.g., extract--from text to json, classification--sentiment classification).

We believe that data quality has a pivotal role in model quality, hence we have taken a systematic approach to ensuring data quality. Data from web is usually skewed in quality, even more so for corpus outside English. On the other hand, the mechanism underlying LLMs makes models have good memory. Our experiments find that less than a dozen low-quality examples would have the model learn and generate suboptimal results. Examples of low-quality data include: casual or oral language on the web, jargons in social media, ads, harmful contents, messy format and style, and so forth. We developed a set of rules and models to filter out about 10\% low-quality data as follows. Given the large volume of the data, we designed the approach to have a complexity of $\mathcal{O}(n^2)$.

\begin{enumerate}
	\item Filter out ill-format data using a set of rules, e.g., too many symbols or digits in prompts.
	\item Dedup using exact string match.
	\item Dedup using sequence simhash $+$ longest common substring.
	\item Filter out harmful content using an ensemble of three SOTA safety models and a dictionary of sensitive words.
\end{enumerate}

\subsection{Training method}
We innovate on the shoulders of our precedents, to fully inherit the data, computational resources, and intellectuality laid upfront. Models with 7, 13, and 70B parameters were continual pretrained from Llama-2 respective variants, whereas 180B was pretrained from BLOOM. We adopt most of the training setting and model architecture from our precedents, mainly decoder-only transformers~\cite{Vaswani:2017aa}, RoPE~\cite{Su:2021aa} and ALiBi~\cite{Press:2021aa} positional embedding, SwiGLU~\cite{Shazeer:2020aa} and GeLU activation functions, respectively. We made further contributions as elaborated below. Our design objectives are: (1) to put forward a SOTA training infrastructure that can yield superior models in a computational economic manner, (2) multilingual coverage especially for Chinese, and (3) instrumental for application development.

\paragraph{Tokenizer}
Llama models have been lacking language representation other than English (e.g., Chinese only accounts for 0.13\% in their training data), hence we expand tokenizer vocabulary in Tigerbot. First, we sample a 100GB corpus of Chinese and major eastern asian languages (mainly Japanese and Korean) from our pretraining data. Second, we train a Byte-Pair Encoding (BPE) SentenePiece tokenizer~\cite{google:2023aa} using this non-English data to ensure representativeness in final tokenizer. Third, we merge the non-English tokenizer with the original Llama-2 one to make our final tokenizer~\cite{Cui:2023aa}. The original vocabulary size is 32k, we expand to near but not exceeding 65k, to avoid doubling storage and IO in downstream tokenized binarized data. We also found that more than 100GB training data is unnecessary, the character coverage is almost identical; but the peak CPU memory would exceed 2TB, which is beyond mainstream hardware. For our 180B-parameter model, we keep the vocabulary size same as 250k, since BLOOM already has good multilingual coverage.

\paragraph{Training framework}
TigerBot model family has been trained using our private fork of Megatron-DeepSpeed~\cite{microsoftP2023aa}, which implements 3D parallelism by combining ZeRO sharding, data parallelism (DP) and pipeline parallelism (PP) from DeepSpeed~\cite{Rajbhandari:2019aa} with tensor parallelism (TP) from Megatron-LM~\cite{Narayanan:2021aa}. Our fork made several upgrades as follows: 

\begin{enumerate}
	\item Bring Megatron-DeepSpeed Modeling class up to speed with SOTA architectural ingredients including CoreAttention, SwiGLU, grouped-query attention (GQA)~\cite{Ainslie:2023aa}, RoPE~\cite{Su:2021aa}, and flash attention~\cite{Dao:2022aa} adapted to Llama-2 architecture.
	\item Design a better yet simple algorithm for pipeline partition. Given that a model with $N$ attention blocks is divided into $M$ stages, first $N\ mod\ M$ stages contain $\left\lceil{N/M}\right\rceil$ blocks and remaining blocks each has $\left\lfloor{N/M}\right\rfloor$ blocks. Compared to the original implementation where total stage is limited to several special numbers, our method is more flexible to alleviate the problem of skewed partition.
	\item A set of scripts converting Megatron-DeepSpeed sharded weights to transformers weights and vice versa.
\end{enumerate}

Tensor parallelism is particularly critical for training models over 100 billion parameters, since model size over 70B cannot fit into a single GPU while CPU offloading is slow and we want to avoid. On the other hand, more TPs introduce heavier communication and TP across nodes is impractical due to huge inter-node communication. Given a fixed value of $\text{TP}\times\text{PP}\times\text{DP}$, empirically we found smaller TP yields better global efficiency, largely because of heavier communications among tensor partitions (parallelizing matrix multiplication) than those of pipeline partitions (layers). Along together with techniques of GQA, flash attention, gradient accumulation and checkpointing, we have been able to find optimal configurations for different model sizes under various cluster resources. A back-of-the-envelop calculation follows.

\begin{figure}[h]
\centerline{
\subfigure[iteration time vs. tokens]{\includegraphics[width=0.29\linewidth]{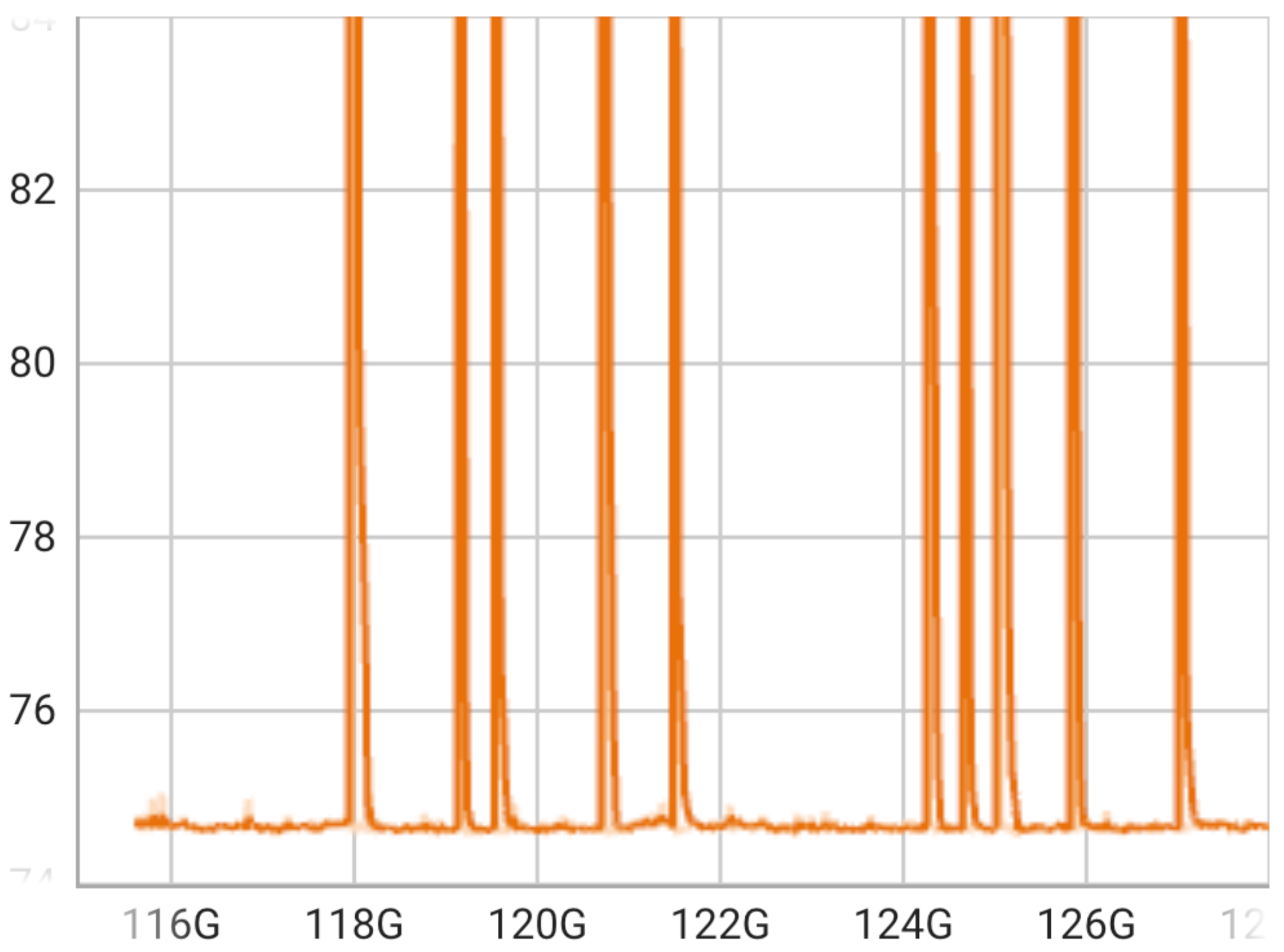}
\label{fig-efficiency-sampsec}}
\subfigure[batch size vs. tokens]{\includegraphics[width=0.3\linewidth]{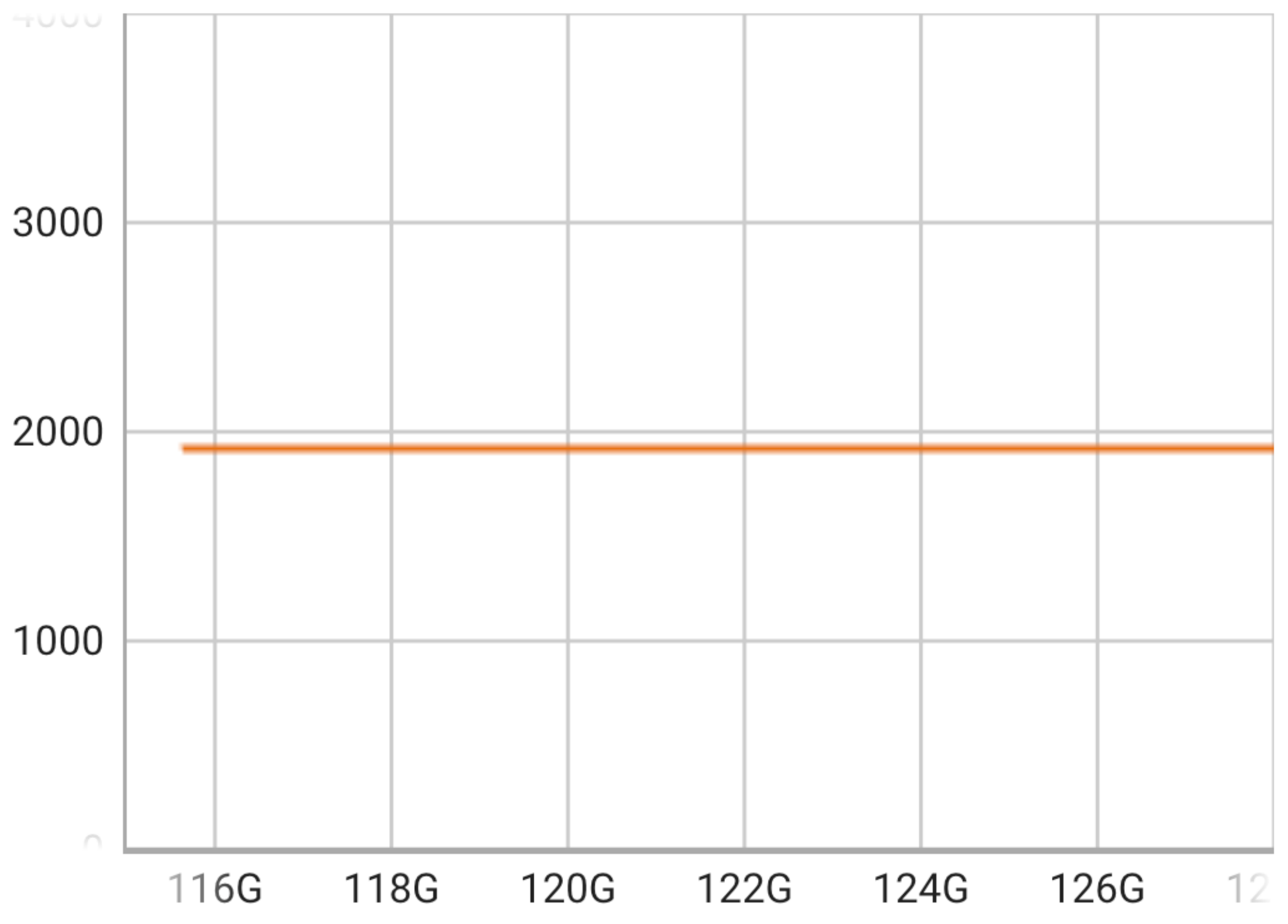}
\label{fig-efficiency-gbs}}
\subfigure[seqlen vs. tokens]{\includegraphics[width=0.3\linewidth]{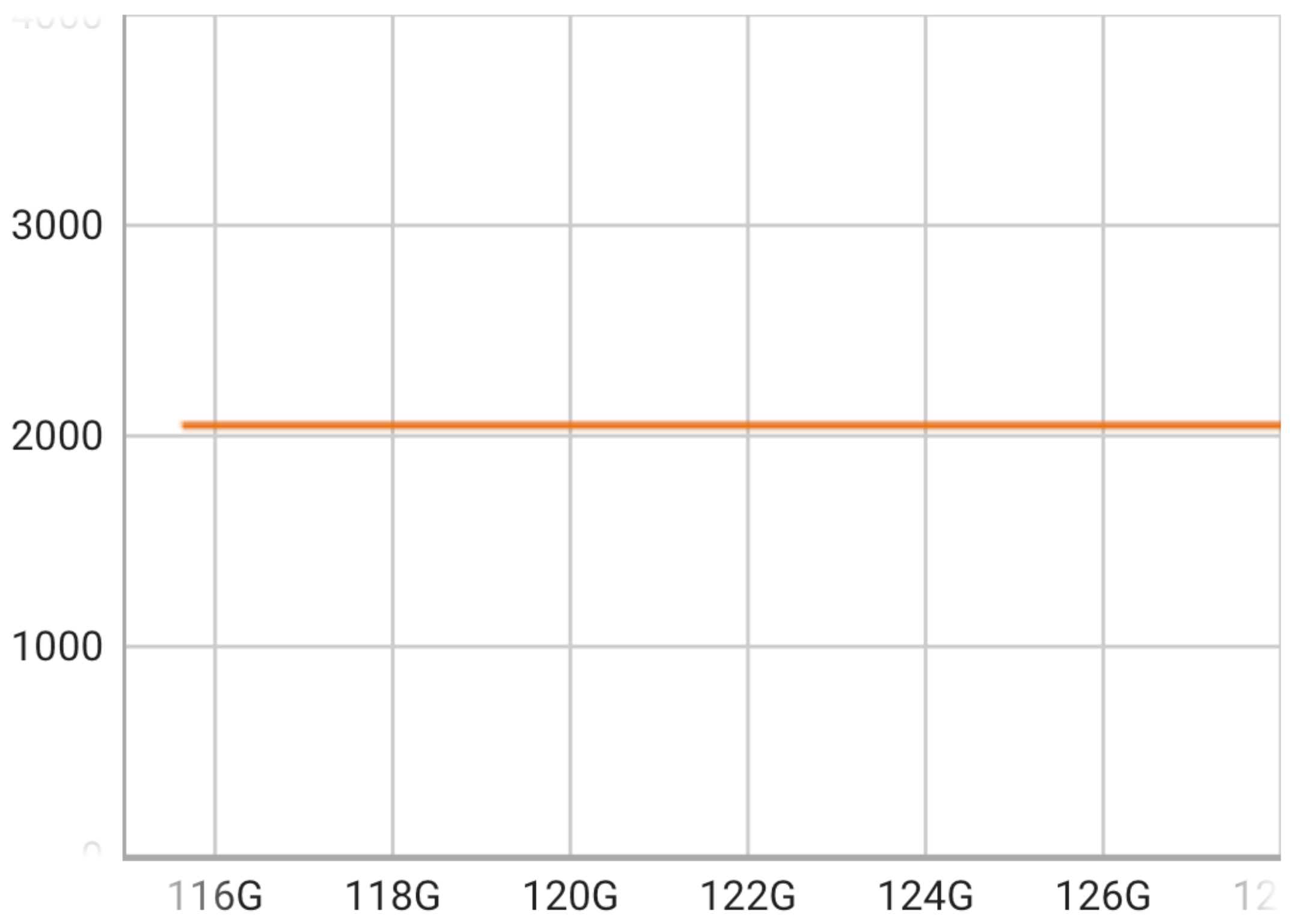}
\label{fig-efficiency-seqlen}}
}
\caption{Training efficiency for Tigerbot models}
\label{fig-efficiency}
\end{figure}

Llama-2-13B pretrain GPU hours is 368,640 with 2TB tokens data, per the Llama-2 paper~\cite{Touvron:2023aa}, thus we have: $\text{training-tokens}/\text{gpu-sec}=1,507$. On the other hand, TigerBot-13B training throughput reaches $25.7~\text{examples}/\text{sec}$ or equivalently $74.6~\text{sec}/\text{iteration}$, as shown in Figure~\ref{fig-efficiency}, on a $32\times$ A100-40G GPU cluster, with a sequence length of $2,048$. Only after a preliminary geometric search of aforementioned parallelism configurations, we found an optimal setting is: TP=2, PP=8, DP=2, per-device-batch-size=2, and global-batch-size=1,920, to reach about 4M tokens global batch. Our efficiency reads: $\text{training-tokens}/\text{gpu-sec}=1,645$ (109\% of Llama-2 training). Also considering that Llama-2 used higher-end Meta’s Research Super Cluster (A100-80G, 2TB CPU memory, RDMA inter-connection)~\cite{Peckham:2023aa}, we believe that TigerBot's codebase has reached cutting-edge computational economics world wide.

\paragraph{Holistic training}
In the pretraining stage, we mix in 2-5\% (of pretraining data) instruction completion data, preprocessed into an unsupervised format, e.g., \verb|{instruction}+"\n"+{response}|, and remove duplicates from the original pretraining data that overlaps with SFT data knowledge. The rationale behind is: instruction completion (e.g., question-answering, generation) is essentially still a kind of human language continuation. We want models to learn some patterns of instruction following, holistically along with base knowledge during pretraining. The incurred additional computational cost is negligible, but the advantages are two folds:

\begin{enumerate}
	\item Base models exhibit strong capability to follow instructions, right away before alignment. We performed a quick evaluation on SQuAD2.0 benchmark, and found that Tigerbot-13b-base reached 86\% of the next token prediction accuracy as Tigerbot-13b-chat.
	\item Since foundational capabilities (knowledge and instruction following) has been learned during pretraining, alignment learning can be lightweight. This further benefits rapid and economic application deployment in various verticals. Our experiments show that loss reaches 95\% of convergence after one million examples of SFT training.
\end{enumerate}

\paragraph{Supervised fine-tuning (SFT)} The models are only trained on the response portion of supervised training data, using largely the same training routine, except data preprocessing. We first introduce a pair of special tokens to markup instruction and response, respectively, for each example. We then process data examples individually (truncate or pad to maximum sequence length) or grouped (concatenate into maximum sequence length) into trainer. For the grouped method, we implement attention mask such that attention would not be computed cross examples. SFT data is usually quite sparse (few long examples and most within \verb|max-seq-length|). The grouped method gives 5 to 10$\times$ speedup, albeit the individual way should yield higher accuracy intuitively. However, our experiments show that there is no noticeable performance difference between two methods, we then choose the grouped method in our production runs. We also notice that grouping method may introduce some degree of robustness and generality into the model, analogous to that human also learn from noises.

\paragraph{Reinforcement learning with human feedback (RLHF)}
In the RLHF stage, we first adopt rejection-sampling fine-tune with human-in-the-loop as follows:

\begin{enumerate}
	\item Sample 10k prompts from SFT data and real user interactions.
	\item Generate 10 random responses (e.g., with temperature=0.6) for each prompt, using our best candidate chat model.
	\item Rank the generations, of all prompts 90\% using reward model (a 13B model fine-tuned for text classification), 5\% using human ranking, and 5\% using human editing (gold set).
	\item Mix the top-ranked generations into SFT data and perform fine-tuning.
	\item Iterate the above process on a monthly basis, as we collect real user prompts.
\end{enumerate}

We further apply the direct preference optimization (DPO) algorithm~\cite{Rafailov:2023aa} on top of the rejection-sampling fine-tuned weights, using 5k gold comparison data. We choose DPO for it is simple to implement, efficient to train, while performing as well as or better than PPO-based methods. By intrinsically treating the LLM to be fit as a reward model, DPO essentially formulates a classification problem for pairwise comparison, with a simple cross-entropy loss as follows~\cite{Rafailov:2023aa}:

\begin{equation}\label{eqn-dpo}
	\mathcal{L}_\text{DPO}(\pi_\theta;\pi_\text{ref})=-\mathbb{E}_{(x,y_w,y_l)\sim\mathcal{D}}\left[\log{\sigma\left(\beta\log{\frac{\pi_\theta(y_w|x)}{\pi_\text{ref}(y_w|x)}}-\beta\log{\frac{\pi_\theta(y_l|x)}{\pi_\text{ref}(y_l|x)}}\right)}\right]
\end{equation}

where $\sigma$ is a logistic function, and $\beta$ is a hyperparameter that controls the deviation from the reference policy $\pi_\text{ref}$. Also as in~\cite{Rafailov:2023aa}, empirically we found DPO is more stable to train, likely due to its elegant formulation and does not need separate networks as in actor-critic style PPO algorithms. Efficient and stable training leads to rapid iteration in application deployment.

\paragraph{Long sequence}
\label{sec-longseq}
Long sequence capability of a model tends to be important in applications, e.g., long context window to read a novel, or long generation length to write a book, all at once. Since text data follows a power law, i.e., most is short and few is long, inference with long sequence can be analogous to magnifying a picture. There are two key factors: (1) the resolution of the original image, and (2) the interpolation algorithm. The former is the length of training samples and the latter is RoPE extrapolation method.

During training TigerBot, we increase the RoPE base frequency to 500k~\cite{Xiong:2023aa} and group training samples to 4k. Attention parallelism could also be done along the sequence length dimension to allow training length exceed the limit of total GPU memory on a single node. This may yield extra-long sequence, e.g., several hundred thousand for certain applications, but at a higher cost of speed. Tensor parallelism between nodes becomes extremely slow due to huge communication, which indirectly limits the sample length in each node. We choose not to mix in special-purpose long-sequence data to preserve the generality of our models, but in the alignment stage we observed about 2\textperthousand\ examples exceeding 2k tokens.

During the inference phase, length extrapolation is achieved through interpolating RoPE position embeddings~\cite{Su:2021aa}. The main difference among popular methods like Dynamic and YaRN~\cite{Peng:2023aa} lies in their interpolation techniques. The challenge is how to maintain output consistency. In the implementations of Dynamic and YaRN by Transformers~\cite{Haggingface:2023aa} and TGI~\cite{Huggingface:2023ab}, the approach involves "caching the position embeddings of the longest sequences seen". Under this implementation, even if the model observes the same input, the output may differ due to variations in the length of the cached position embeddings. Tigerbot addresses this by calculating the sum of the \verb|input-token-length| and \verb|max-new-token-length| per request. This value is used as a reference when computing scaled position embeddings. This ensures the model's performance remains consistent when extrapolating lengths. The model's behavior for sequences not exceeding the training length is also preserved. We extrapolate the max sequence length to 32k using a RoPE scaling factor of 8.

\paragraph{Quantization}
Quantizing a LLM involves using a reduced-precision integer representation for weights and activations, which might be important for practical considerations of GPU memory limitation and fast inference. We implemented both static and dynamic quantizations.

In static quantization, the weights and activations of the model are computed using a calibration dataset in advanced. TigerBot models are quantized using ExLlamaV2~\cite{Turboderp:2023}, which is based on the same optimization method as GPTQ. We demonstrated up to 3$\times$ speedup and 4$\times$ memory reduction for Tigerbot-4bit quantized models with negligible loss in accuracy.

In dynamic quantization, the weights are still quantized ahead of time post training, but the activations are quantized during inference on the fly. In particular, we use 8-bit weight and 16-bit activation quantization (W8A16). Our experiments show that 8-bit activation may incur significant accuracy degradation similar to~\cite{Yao:2022aa}, while W8A16 yields a good balance between accuracy and speedup. The dynamic approach has advantages in adapting to various serving hardwares, especially for those bottlenecked more by memory bandwidth than compute.

\paragraph{Safety}
We have performed safety filtering on the training data, we also take measures to mitigate safety risk during training and inference at runtime for any user-interfacing TigerBot products. We adopt a safety category consisting of 5 categories and 31 subcategories. Main categories include:

\begin{enumerate}
	\item Violating core values of national and social security
	\item Discriminatory content
	\item Commercial illegal and unregulated activities
	\item Infringement of others' legitimate rights and interests
	\item Inability to meet the security requirements for some special-purpose services, e.g., medical information services and critical information infrastructure.
\end{enumerate}

During training, we use human annotation to collect about 40k safety demonstration data, in consultation with administration guidelines and domain experts. This data is then fed into our alignment learning as well as pretraining per holistic training. The safety training data is refreshed on a monthly basis and reflected into our iterative alignment process. Both data and training level safety measures are preventive, while runtime-level safety check is protective.

During runtime inference, user input is examined safety first before feeding into model to generate. Would either user input or model generation be detected harmful, our products provide a default yet suggestive response to users. All contents undergo a two-stage examination, first a dictionary of about 120k sensitive vocabulary, followed by an ensemble of three BERT-based classifiers. These safety classifiers are trained on millions of annotated positive (violating) samples and focused on different aspects in the aforementioned safety categories. The dictionary is designed to be comprehensive to ensure high recall, whereas a good precision is achieved by tuning the positive threshold from safety classifiers. The final safety label is a parameterized function of dictionary detection and classifier output, and may well be varied for different domains and applications. Moreover, we have safety team to keep our dictionary and classifiers up to date with emerging topics and administration guidelines.

\begin{figure}[h]
\centerline{
\subfigure[training loss]{\includegraphics[width=0.4\linewidth]{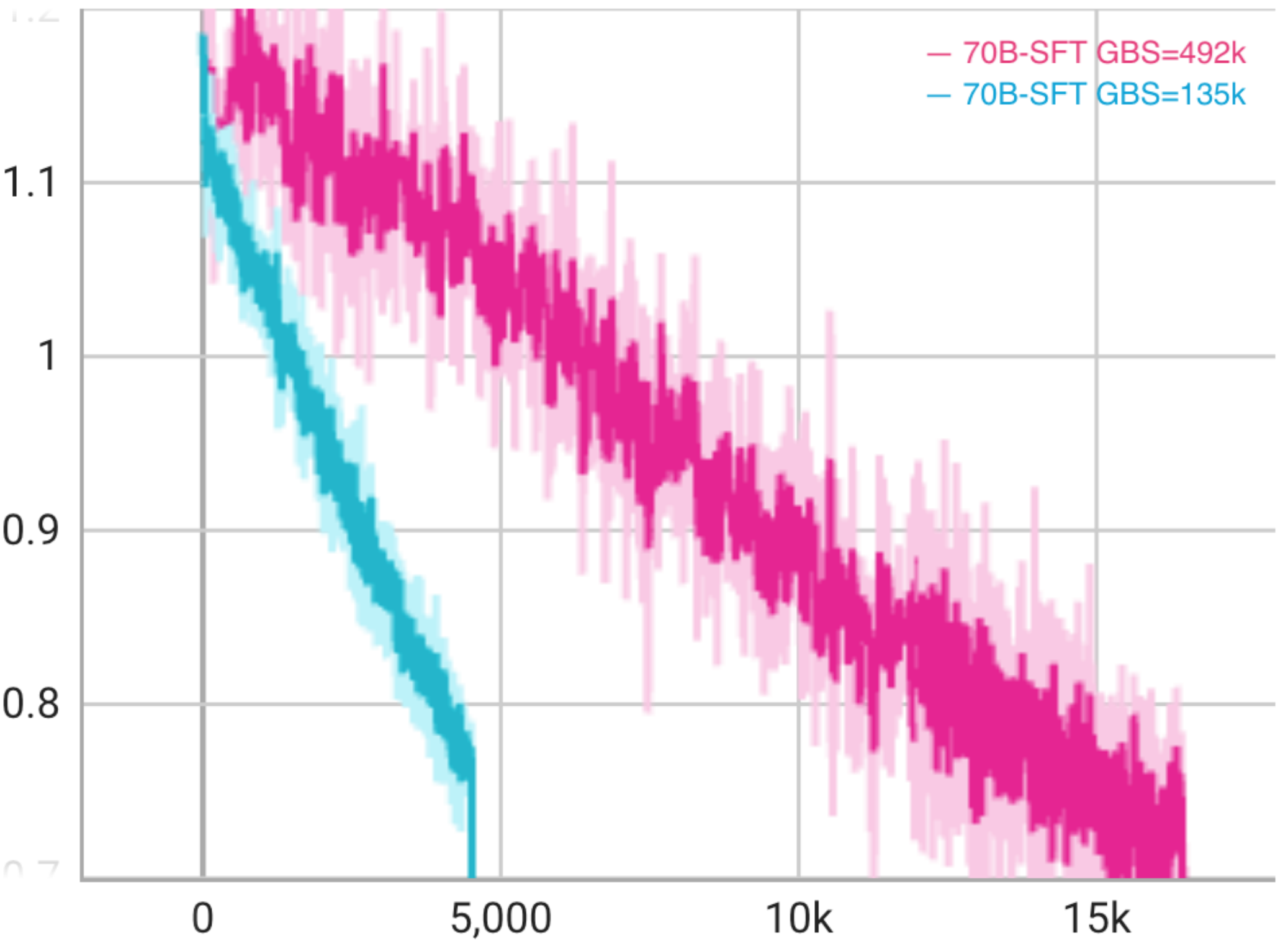}
\label{fig-loss-train-sft}}
\subfigure[validation loss]{\includegraphics[width=0.405\linewidth]{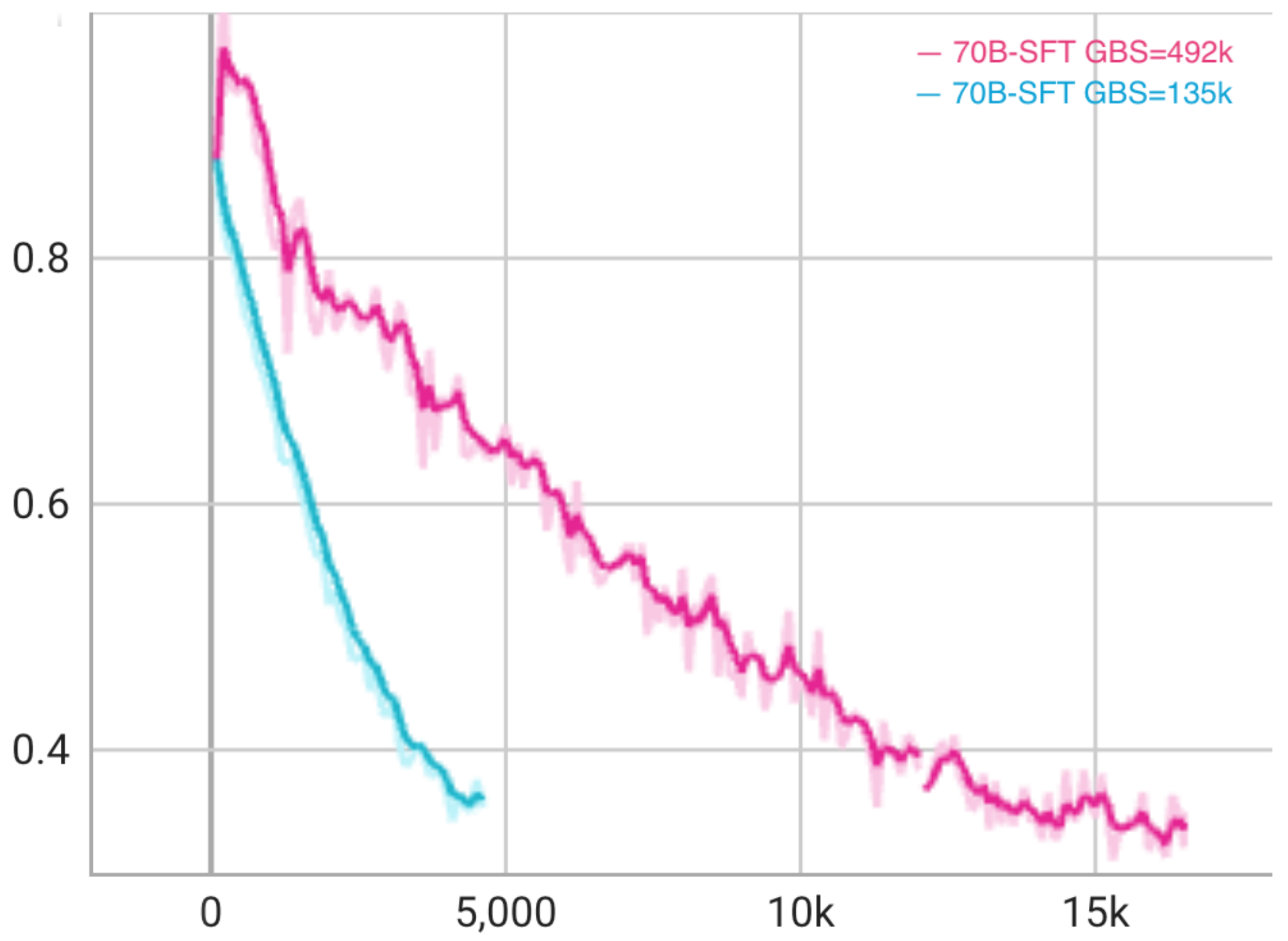}
\label{fig-loss-valid-sft}}
}
\caption{Tigerbot-70b-SFT loss with different batch size}
\label{fig-loss-sft}
\end{figure}

\paragraph{Hyperparameters}
We pretrained TigerBot models using a global batch size (GBS) of 4M tokens, while fine-tuned models with a GBS as small as 100--400k tokens. Our experiments show that, given high quality data, smaller batch for finer-grained updates can yield lower loss, as shown in Figure~\ref{fig-loss-sft}. We pretrained models for one epoch over training data, fine-tuned for two epochs, then followed by alignment learning for one epoch.

We used the adamW optimizer, with $\beta_1=0.9, \beta_2=0.95, \epsilon=10^{-5}$. We applied a cosine decaying learning rate (LR) schedule between $[2.0^{-5}, 2.0^{-4}]$ for 7B and 13B, and $[1.0^{-5}, 1.0^{-4}]$ for 70B and 180B model pretraining. For fine-tuning, the LR is $[2.0^{-6}, 2.0^{-5}]$. We used warmup steps close to 1\% of training steps, a weight decay rate of 0.1, and gradient clipping to 1.0. All trainings were performed under bfloat16 precision and flash attention, except for 180B we used gradient checkpointing instead.

\paragraph{Training hardware}
Our training cluster consists of $512\times$ A100-40G GPUs (64 nodes $\times$ 8 GPUs), equipped with NVLink intra-node GPU interconnects, and RoCE (RDMA over Converged Ethernet) inter-node communications. Each node has 64-core Intel Platinum CPU and 1024GB of RAM.

\subsection{Evaluation}
Evaluating LLMs is a challenging yet critical problem. It serves as a systematic basis for model selection, and reveals areas for improvements. However, performing evaluation only at the end of the training pipeline puts expensive computations at a risk of sinking, possibly just for a small negligence in data or bugs in codes. Moreover, developers have found that often there are discrepancies between benchmark metrics and human perception. To address these challenges, we develop a three-stage evaluation approach as below.

\begin{table}[h]
  \caption{Tigerbot base model evaluation results}
  \label{tab-eval-base}
  \centering
  \begin{tabular}{cclclclc|c|c}
    \toprule
    \multirow{2}{*}{Lang.} & \multirow{2}{*}{Task} & \multirow{2}{*}{Benchmark} & \multicolumn{2}{c}{TigerBot} & \multicolumn{2}{c}{Llama-2} \\
    \cmidrule(r){4-5}\cmidrule(r){6-7}
    &&& 70B-base & 13B-base & 70B-base & 13B-base \\
    \midrule
     \multirow{9}{*}{En} & Code & HumanEval & 28.66 & 18.29 & 31.10 & 15.27 \\
     \cmidrule(r){2-3}
     &\multirow{5}{*}{\makecell{Commonsense\\Reasoning}} & PIQA & 83.30 & 79.33 & 82.21 & 79.21 \\
     && SIQA & 48.77 & 48.52 & 46.01 & 46.32 \\
     && HellaSwag & 78.62 & 72.64 & 79.46 & 74.96 \\
     && WinoGande & 69.38 & 64.17 & 69.69 & 64.09 \\
     && OpenBookQ & 88.60 & 72.40 & 57.40 & 57.00 \\
     \cmidrule(r){2-3}
     &\makecell{Reading\\Comprehension} & BoolQ & 67.52 & 63.18 & 69.69 & 71.50 \\
     \cmidrule(r){2-3}
     & Math & GSM8K & 65.66 & 35.86 & 63.99 & 28.81  \\
     \cmidrule(r){2-3}
     &\makecell{Multi-choice\\Questions} & MMLU & 68.68 & 55.35 & 69.58 & 55.81 \\
     \cmidrule(r){2-7}
     \multicolumn{3}{c}{\textbf{Average (En)}} & \textbf{66.58} & \textbf{56.64} & \textbf{63.24} & \textbf{54.77} \\
     \midrule
     \multirow{4}{*}{Zh} & \multirow{2}{*}{\makecell{Reading\\Comprehension}} & CMRC & 85.93 & 66.57 & 68.97 & 76.73  \\
     && C3 & 77.37 & 67.01 & 60.16 & 47.51 \\
     \cmidrule(r){2-3}
     &\makecell{Natural Language\\Inference}& OCNLI & 30.00 & 30.60 & 30.03 & 30.00 \\
     \cmidrule(r){2-3}
     &\makecell{Multi-choice\\Questions} & C-EVAL & 67.75 & 48.46 & 49.90 & 38.67 \\
     \cmidrule(r){2-7}
     \multicolumn{3}{c}{\textbf{Average (Zh)}} & \textbf{65.26} & \textbf{53.16} & \textbf{52.27} & \textbf{48.23} \\
     \bottomrule
  \end{tabular}
\end{table}

\begin{table}[h]
  \caption{Tigerbot chat model evaluation results}
  \label{tab-eval-chat}
  \centering
  \begin{tabular}{cclclclc|c|c}
    \toprule
    \multirow{2}{*}{Lang.} & \multirow{2}{*}{Task} & \multirow{2}{*}{Benchmark} & \multicolumn{2}{c}{TigerBot} & \multicolumn{2}{c}{Llama-2} \\
    \cmidrule(r){4-5}\cmidrule(r){6-7}
    &&& 70B-chat & 13B-chat & 70B-chat & 13B-chat \\
     \midrule
     \multirow{10}{*}{En} & Code & HumanEval & 31.10 & 26.83 & 26.22 & 11.59 \\
     \cmidrule(lr){2-3}
     &\multirow{5}{*}{\makecell{Commonsense\\Reasoning}} & PIQA & 83.57 & 80.09 & 80.25 & 78.67 \\
     && SIQA & 51.89 & 49.44 & 51.59 & 50.97 \\
     && HellaSwag & 76.68 & 70.63 & 77.63 & 74.71 \\
     && WinoGande & 67.01 & 63.30 & 68.11 & 65.82 \\
     && OpenBookQ & 85.00 & 67.40 & 85.00 & 80.00 \\
     \cmidrule(r){2-3}
     &\makecell{Reading\\Comprehension} & BoolQ & 80.67 & 78.87 & 78.00 & 78.32 \\
     \cmidrule(r){2-3}
     & Math & GSM8K & 84.91 & 51.25 & 58.91 & 54.62  \\
     \cmidrule(r){2-3}
     &\makecell{Multi-choice\\Questions} & MMLU & 68.03 & 55.94 & 64.84 & 54.61 \\
     \cmidrule(r){2-7}
     \multicolumn{3}{c}{\textbf{Average (En)}} & \textbf{69.87} & \textbf{60.42} & \textbf{65.62} & \textbf{59.43} \\
     \midrule
     \multirow{4}{*}{Zh} & \multirow{2}{*}{\makecell{Reading\\Comprehension}} & CMRC & 85.37 & 76.17 & 80.06 & 74.62  \\
     && C3 & 75.34 & 69.42 & 54.85 & 51.01  \\
     \cmidrule(r){2-3}
     &\makecell{Natural Language\\Inference}& OCNLI & 38.07 & 40.17 & 36.23 & 30.00 \\
     \cmidrule(r){2-3}
     &\makecell{Multi-choice\\Questions} & C-EVAL & 60.40 & 48.89 & 44.17 & 39.22 \\
     \cmidrule(r){2-7}
    \multicolumn{3}{c}{\textbf{Average (Zh)}} & \textbf{64.80} & \textbf{58.66} & \textbf{53.83} & \textbf{48.71} \\
    \bottomrule
  \end{tabular}
\end{table}

\begin{enumerate}
	\item During training for major checkpoints, we perform lightweight evaluations for a quick preview. We first draw a random sample of 100k train and validation examples from 10 major benchmarks including ARC~\cite{Clark:2018aa}, CommonsenseQA~\cite{Talmor:2018aa}, SQuAD 2.0~\cite{Rajpurkar:2018aa}, WebQuestions~\cite{Berant:2013aa}, and so forth. We then implemented a next-token prediction accuracy in Transformers' trainer, and a run of evaluation only takes several minutes, on one node, even for 70B and 180B models.
	\item After training, we conduct a comprehensive evaluation on 13 mainstream benchmarks, covering 8 tasks and languages mainly of English and Chinese. The evaluation datasets are designed to cover a broad range of tasks such as math, reasoning, codes, and reading comprehension, and so forth. We report the results for both base and chat models in Tables~\ref{tab-eval-base} and~\ref{tab-eval-chat}, respectively. Base models are tested using 0-shot, and otherwise we follow the implementation of OpenCompass~\cite{oc:2023aa} to promote reproducibility.
	\item Furthermore, we carry out human evaluations on our top candidates. Human evaluation is usually considered gold standard for assessing natural language generation systems. However, since LLMs can perform a very wide range of tasks, it is impractical to collect a comprehensive dataset to yield statistically significant results. The subjective biases among human annotators and the nontrivial cost if evaluating iteratively are also important considerations. We employ human evaluation as a gatekeeper, in conjunction with automated benchmark to select models. We first collect 5k gold set prompts, largely from real user questions and unseen from any upstream process. The gold set is sampled to cover a broad range of tasks as described in Section~\ref{sec-data}, yet reflect real user behaviors (tones, typos, oral vocabulary, etc.). We then ask our human annotators to rate on helpfulness and safety using a 1-5 Likert scale. We develop an evaluation guideline to train our annotators, which includes detailed rules and examples for various types of generations. e.g., factuality for objective questions, diversity for subjective generations. We occasionally found that human evaluation results were not aligned tightly with benchmark metrics, we choose production models as a holistic tradeoff, also considering model degradation, serendipity and downstream application desiderata.
\end{enumerate}

\section{Applications}
In this section, we elaborate on our implementations of a spectrum of applications, some are tools and consumer products, and some are real world deployed applications.

\paragraph{Long-context question answering}
Many applications involve reading comprehension (summarization and question-answering) based on a long context, e.g., reading papers, studying laws, QA based on an in-house knowledge base, and so forth. Most knowledge-based QA systems are implemented as a two-stage retrieval-reader pipeline. A dense retrieval narrows the context down to a small subset of passages, followed by a LLM reader to generate answers based on the retrieved contexts~\cite{Karpukhin:2020aa}. This setup has two limitations: (1) it is not suitable for summarizing and inductive questions, where context should be comprehensive; and (2) a large portion of errors may already occur at the retrieval stage, due to practical reasons e.g., noisy and ill-format data.

As described in Section~\ref{sec-longseq}, we have extrapolated the context length to 32k tokens, or approximately 50k characters as in a 50-page pdf or word document (Tigerbot's tokenizer has a character-to-token compression rate of 1.5$\times$ for Chineses and 5$\times$ for English). This context window is big enough for most ad-hoc knowledge-based QA tasks, therefore we skip the dense retrieval part and take a one-stop approach as follows.

\begin{enumerate}
	\item Segmentation: if input text exceeds the max input length (e.g., 32k), segment it into chunks of max length, using line breaks to preserve semantics.
	\item Filtering: upon receiving a user query, zero-shot prompt the LLM as a binary classifier to filter out irrelevant segments, similar to the approach in~\cite{Chen:2023aa}. We compose the prompt as: \verb|C:{context}+"\n"+Q:{query}+"\n"+"Can the above Q be answered by C?"|
	\item Generation: first generate a response by each candidate segment as intermediate results, from which then generate the final response using a prompt like: \verb|{intermediate results}+"\n"+{query}|. This recursive approach has been used in summarization task and shown superior performance~\cite{Wu:2021aa}.
\end{enumerate}

\paragraph{Recursive summarization}
Summarization has been one major NLP task, and now can be seamlessly solved by LLMs. To effectively handle extensively long texts, we employ a recursive summarization approach, similar to~\cite{Wu:2021aa} but we solely rely on LLM. We first chunk the input text into smaller and manageable segments (within \verb|max-input-length|), following natural semantic boundaries such as section endings and line breaks. We then summarize each segment independently. In the final step, we aggregate these individual summaries, to generate a comprehensive and cohesive final summary. Domain specifics and desired length can be naturally guided by prompts, e.g., with prompt: \verb|"Summarize the above article into 200 words, preserving key financials."|

\paragraph{Function calling}
Natural language interface is attractive to many applications, where users give instructions in natural language and systems can understand and perform tasks otherwise requiring structured inputs. The intrinsic capability of natural language understanding of LLMs can be leveraged to extract structured data from natural language query, and then to perform downstream tasks, namely function calling. We design the function calling capabilities relying on the underlying TigerBot models, and as three steps as follows.

\begin{enumerate}
	\item Extraction: given a function definition, the underlying LLM is prompted to extract function arguments from a user query. We compose the prompt as: \verb|F:{func}+"\n"+Q:{query}+"\n"+"Extract args from Q per F."+"\n"+JSON:|.
	\item Calling: we then call the target function with the arguments extracted to get a response. Functions can be in-house systems or third-party APIs from web, e.g., stock quotes and weather lookup APIs.
	\item Generation: with the returned function response, we prompt the LLM again as: \verb|{func response}+"\n"+{query}| to get the final answer in natural language.
\end{enumerate}

The end-to-end performance of function calling largely relies on the LLM's capabilities in natural language understanding and structural extraction. For this purpose, we intentionally mixed a mild portion of general-purpose extraction data in pretraining and fine-tuning. We have observed quite satisfactory performance for some basic function calling tasks, e.g., math calculation and stock quotes lookup. Therefore, we believe that with further fine-tuning on domain-specific extraction data, the function calling capabilities can be of use in many real world applications, particularly in place of those costly and complex legacy systems just for structural extraction.

\paragraph{Online search}
LLMs can be augmented by search, to get factoid and real time context, also to some degree to alleviate the hallucination problem. We implement search augmentation as follows.

\begin{enumerate}
	\item Preprocess and search: we first preprocess user query to be suitable for modern search engines, e.g., removing oral interjections and time reference resolution, then issue the query to search engines to get results.
	\item Quality filtering and parsing: we then filter the search results into top 1-3 candidates based on some quality and timeliness conditions, e.g., site quality and if results within a week present, remove those older than one month. We also parse relevant contents from those structured results, e.g., weather and stock prices.
	\item Generation: we finally generate the response to users by prompting the underlying LLM with: \verb|{top search results}+"\n"+{query}|.
\end{enumerate}

\begin{figure}[h]
\centering
\includegraphics[width=0.8\linewidth]{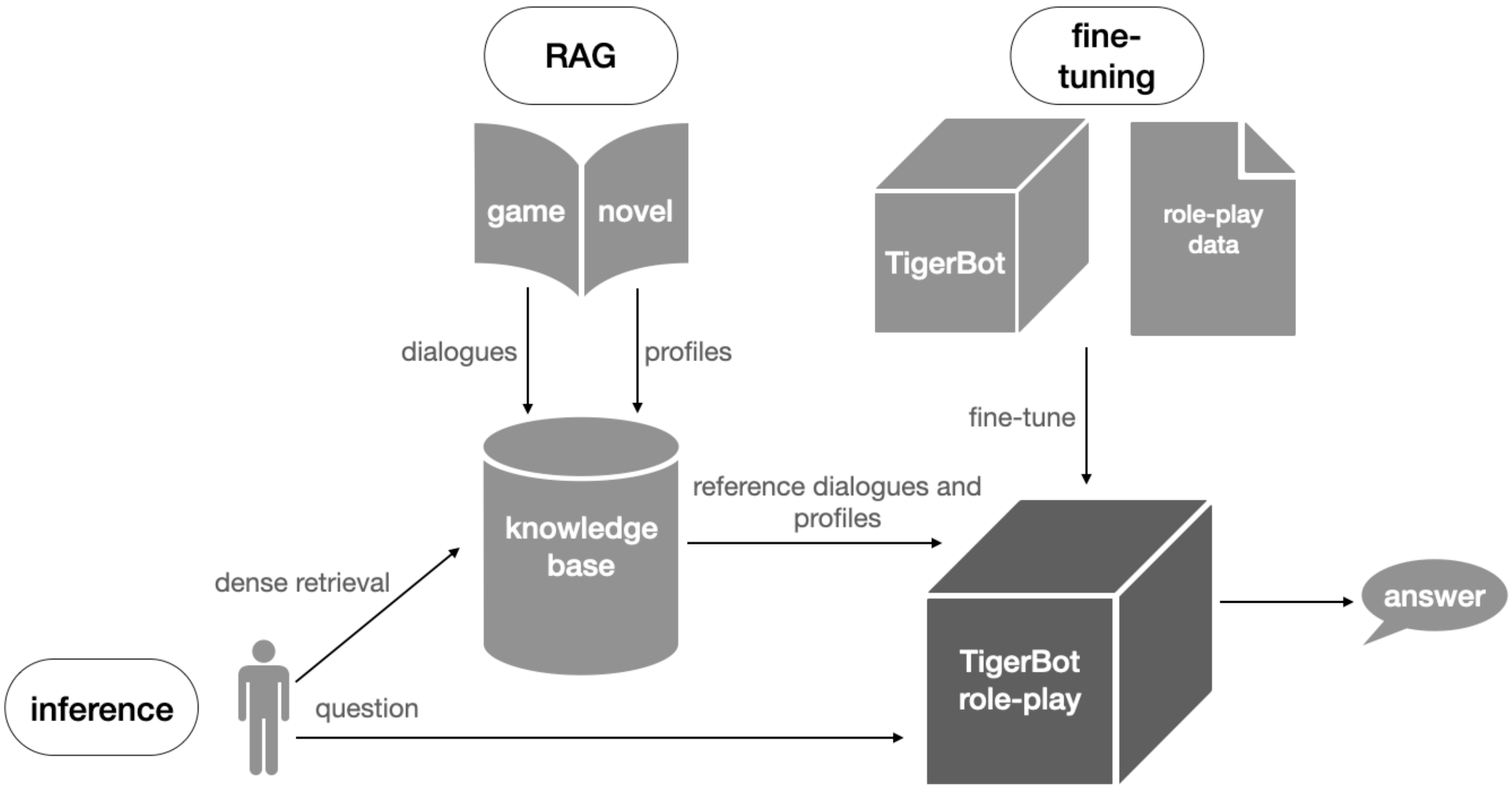}
\caption{A scalable approach to Tigerbot's role-playing LLM}
\label{fig-roleplay-flow}
\end{figure}

\paragraph{Role playing}
Given its rich embodiment of general knowledge and conversational capability, NPC (non-player character) in RPG games can be equipped with LLM to become more entertaining. One common requirement from gaming applications is to make LLMs act as some roles, e.g., dialogues and personal memory. To develop a role-playing LLM, there are two design objectives: (1) to train a LLM with role-playing capability, and (2) to adapt the LLM into the context of the game. Meanwhile, developers usually want to keep the general capabilities of the LLM to make the NPC more humanlike. Furthermore, in practice the approach needs to be somewhat certain, lightweight, and scalable. Our approach to role-playing LLM combines fine-tuning and retrieval-augmented generation (RAG) as follows, and the process is illustrated in Figure~\ref{fig-roleplay-flow}. Our approach was inspired by~\cite{Li:2023aa}, but we added a fine-tuning step to get a gamified foundation.  

\begin{enumerate}
	\item Fine-tuning: we continual fine-tune a TigerBot chat model on a dataset of general-purpose multi-round role-playing dialogues, e.g., acting as a hero in a novel. Fine-tuning can gauge the model to have role-playing capability, albeit is time consuming and stochastic in nature.
	\item Extraction: given a novel or plot as the background context of a game, we extract dialogues between characters and summarize their profiles, both using a general-purpose TigerBot chat model. Extracted dialogues and profiles are fed into an embedding index as the game knowledge base, under a hierarchy beginning with "role".
	\item Inference: during runtime inference, given a user's role and question, we first dense-retrieve reference dialogues and profiles from the knowledge base, and then use the above fine-tuned role-playing LLM to generate response. The prompt is composed as: \verb|{role profiles}+"\n"+{reference dialogues}+"\n"+{question}|.
\end{enumerate}

\begin{figure}[h]
\centering
\includegraphics[width=1\linewidth]{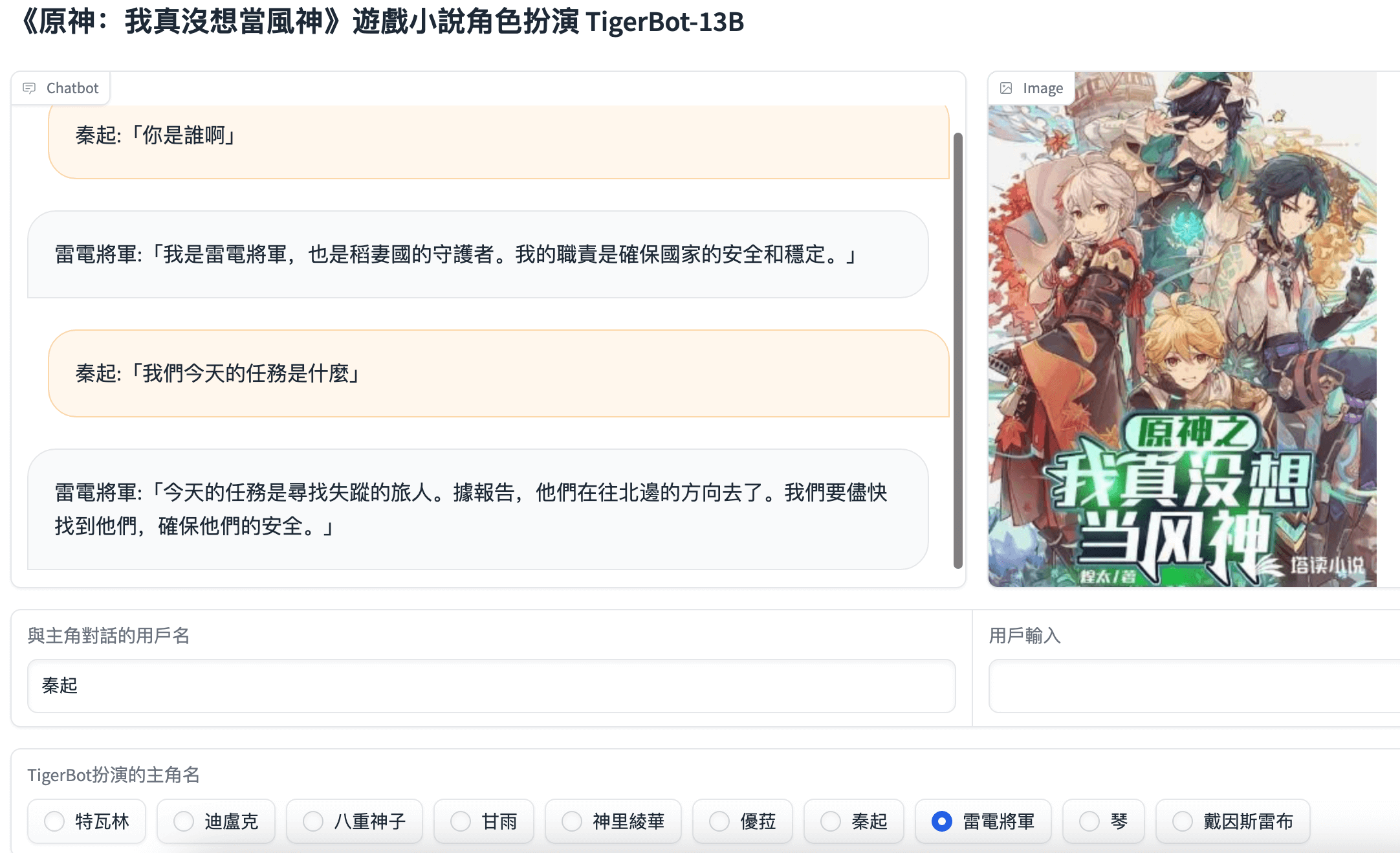}
\caption{An example screenshot of a Tigerbot's role-playing with the novel "Genshin Impact"}
\label{fig-roleplay-eg}
\end{figure}

Our approach is designed such that it can be scaled to different games rapidly and economically. The role-playing capability is general and learned through fine-tuning, while knowledge base extraction is specific to game and fast. An example shown in Figure~\ref{fig-roleplay-eg} is from a role-playing game developed within 2-3 days and took one node of 8$\times$ A100 GPUs.

\paragraph{Intelligent hardware}
Intelligent hardwares can be equipped with LLMs to first have natural language interfaces and potentially to orchestrate other applications using capabilities like function calling. In our practice at this stage, there are three typical requirements as follows.

\begin{enumerate}
	\item Instantaneous response: intelligent hardwares, such as speakers, wearables, and electric vehicles, are mainly used on-the-go, thus require response time to be near real time. The system end-to-end response time usually cannot exceed 2s, while there are other processes involved including ASR, TTS, etc. To achieve fast response, we use streaming generation, along with our optimized inference engine TGI (vLLM and KV cache) as described in Section~\ref{sec-longseq}.
	\item Frequently asked questions (FAQ): intelligent hardwares are purposed for specific scenarios. To get users familiar with the setting, LLMs need to be able to answer a set of FAQs, e.g., What are the make and model of the car?. We first continual fine-tune a TigerBot chat model on domain data, e.g., user manual. Then during inference, we generate answers augmented by retrieval from a knowledge base filled with annotated FAQs. The knowledge base can be easily kept up to date.
	\item Personification: hardwares become more vivid and entertaining would have been portrayed as a real-world character. Personification is manifested by a set of annotated data describing her profile (name and gender) and style (fun or professional). We then fine-tune a TigerBot chat model on the personification dataset for one to two epochs. To preserve the original general-purpose capabilities, the fine-tuning data is mixed with some general data. One rule of thumb for the mixture rate is that general data accounts for more than half.
\end{enumerate}

\section{Conclusion}
In this work, we have introduced TigerBot, a family of pretrained and chat LLMs with parameter sizes of 7B to 180B. TigerBot has achieved SOTA performance with its high-quality training data and a stack of cutting-edge training methods and systems. We also placed an extra emphasis on practical applications, with our detailed implementations and observations for a spectrum of tools and real-world scenarios. We are strong believer of open innovations, and our work has benefited hugely from the LLM open-source community. Likewise, we hope that our work can contribute to the community for future theoretical and practical research and development.

The emergence of LLM has marked one of the most heart-bumping moments in decades of AI development history, largely in view of its overwhelming performance with extreme generality while being straightforward to build. We feel no less awe-inspiring. From our tremendous amount of experiments and implementations, however, we believe that we are still in the nascent stage of LLM and more broadly AGI evolution. There are challenges in reliable solutions to mission-critical tasks, sustainable infrastructure, disruptive not incremental user values from practical applications, to name just a few. The journey ahead is exciting yet equally arduous. Stay calm and happy coding.

\small{
\bibliographystyle{abbrv}
\bibliography{tigerbot}
}


\end{document}